\documentclass[a4paper]{svproc}
\usepackage{cite}
\usepackage{float}
\usepackage{amsmath}
\usepackage{url}
\usepackage{booktabs}
\usepackage{graphicx}
\usepackage{subcaption}
\usepackage{xcolor}         
\usepackage{pifont}

\usepackage[dvipsnames]{xcolor}
\definecolor{darkgreen}{rgb}{0,0.5,0}
\newcommand{\greencheck}{\textcolor{darkgreen}{\ding{51}}}
\newcommand{\redx}{\textcolor{red}{\ding{55}}}

\begin{document}
\mainmatter              
\title{Deploying Foundation Models for Embodied Navigation}
\titlerunning{FMs for Embodied Navigation}  
%
\author{Vishnu Sashank Dorbala\inst{1} \and Dinesh Manocha\inst{1}
}
\authorrunning{Dorbala et al.} 
\institute{University of Maryland, College Park\\
}

\maketitle              

\begin{abstract}
We present and tackle two problems associated with deploying Foundation Models (FMs) on Embodied Agents performing \textit{navigation}: 1) Training bias in FMs leading to poor personalization in unseen environments, and 2) Limited FM context length hindering success, especially on long horizon tasks. Our solution for the former involves priming the FM with human-habit data mined from the scene and our solution for the latter involves active memory management via a novel `memory head' augmentation.
We first present a taxonomy of existing literature on FM-based Embodied Navigation, and highlight these limitations. We then present our approaches, \textbf{Transit-Aware Planning (TAP)} and \textbf{MemCtrl} to address the limitations.
With TAP, we present real-world results in a lab environment with a Turtlebot for personalized target finding that shows an average improvement of $18\%$ over a non-TAP baseline.
On MemCtrl, we report a $6\%$ average improvement across various embodied tasks, with $20\%$ on long instruction subsets, all while using nearly half the context used in the baseline model. Motivated by these result, we present our stance the deployability of FM-based embodied agents in real-world environments, and highlight open research directions.

\end{abstract}

\section{Introduction}

Embodied (or Physical) AI agents interact with the world around them to achieve objectives described by humans. The descriptions of these objectives could be multi-modal, as wayfinding instruction text \cite{r2r, qiREVERIERemoteEmbodied2020}, images describing target views \cite{ignav1, imgnav2,imagenavtopo}, or even vocalizations of sounds \cite{vocalsound, audionav2} in the environment. Tasks for these agents are often described according to what they could expect to encounter in the real-world, be it finding an object, folding clothes, or cooking a meal. The ultimate goal for a generalist agent is to perform a wide range of tasks ``out-of-the-box'', with minimal to no human effort in its decision making.

A common paradigm to developing such generalist agents is to train and evaluate them in simulation environments that aim to capture real-world complexities. Inferring these simulator-trained models on real-world robot agents however assumes comparable on-device compute to that of a simulator setup, which may often not be the case. This compute challenge becomes more apparent while deploying large billion-parameter Foundation Models (FMs). While using FMs has helped improve generalist capabilities, they are limited by fixed context size and poor personalization to novel environments.


In this work, we first present a brief taxonomy of FM-based approaches for Embodied Navigation. We discuss limitations of these approaches in relation to zero-shot performance and deployment. We then present two novel approaches, \textbf{Transit Aware Planning (TAP)} and \textbf{MemCtrl}, which specifically aim to address training and inference shortcomings of FM-based approaches on Embodied agents.
TAP addresses the issue of personalized path planning in dynamic scenes, while MemCtrl introduces a novel memory based architecture that adds a memory head on top of an FM backbone for improved memory efficiency.
Both these approaches aim to improve the deployability of Embodied agents in zero-shot settings, i.e., new environment without any finetuning.
Motivated by these improvements, we also present a few promising directions for research towards enabling the deployment of embodied agents in human-centric environments.

\section{Background}


Navigation is a fundamental capability of embodied agents, and is the main objective of several works in literature including Image-Goal Navigation \cite{imgnav1,imgnav2,imgnav3,imagenavtopo}, Vision-and-Language Navigation \cite{vlnbert,qiREVERIERemoteEmbodied2020,clipnav,vlnog}, and ObjectNav\cite{objectnavrevisited,objg1,objg2,objg3,objg5,lgx}. While a lot of prior classical approaches exist for such language-guided tasks \cite{class1, class2, class3}, our focus for this work is on foundation model based methods. These tasks assume that an agent must first learn to follow guidance in an unseen, unknown environment (`\textit{find a mug}') to get to an appropriate target location, before it can carry out the required manipulation actions (`\textit{pour coffee in it.}'). This becomes especially challenging when agent has no map of the environment, and/or is given minimal guidance; both of which are reasonable assumptions to make for when the agent is deployed in the real-world. 
An example of this would be unboxing a robot in a new home that it has never seen before and asking it to find your shoes. While this seems deceptively simple, it is a complex task that requires the agent to make decisions about navigating the environment with absolutely no knowledge of the layout or structure, while also needing to identify the target object from the guidance and visually ground it in the scene.

\noindent A similar vein of work exists with manipulation on embodied agents, in the form of tasks like Vision-and-Language Manipulation \cite{zheng2022vlmbench, vlmopen, vlm3, manip3}, Pick and Place \cite{pickplace1,uncons1,pickplace2}, Deformable Object Manipulation \cite{defo1, defo2}, and Image-Goal Manipulation \cite{igmanip1,igmanip2}. Tasks like Object Rearrangement \cite{batra2020rearrangement} require the navigating to and manipulating a sequence of several objects for success. Performing these tasks in unseen settings poses a different set of challenges from those in navigation, such as estimating material properties before grasping and accurately modelling contact during interaction.

\noindent Many early learning-based approaches introduced for these tasks involved fine-tuning Imitation Learning (IL) or Reinforcement Learning (RL) models with observations gathered from the environment to improve performance \cite{pirlnav}. While this works well in simulators, IL/RL models face issues with sampling inefficiencies during a real-world transfer, ofen exhibiting poor \textit{zero-shot} performance. 

\noindent More recent methods use Foundation Models (FMs), a type of very large deep learning model that is trained on internet-scale data to learn useful representations for highly improved performance on several downstream tasks. FMs are used either as part of a larger robot planning system, or as an end-to-end scheme, and have shown large performance improvements in zero-shot settings. They however face fundamental issues in zero-shot embodied deployment when the task-relevant context needed for action is not already present in the FMs inputs, latent representation, or the agent’s observation history. In addition, they also face issues related to limited context length, poor interpretability, and memory inefficient architectures that hinder deployment. Embodied agents, just like any other new technology, must exhibit a high degree of Perceived Usefulness \cite{tam} when deployed in the real-world, and achieving good zero-shot performance is a core part of this.

\section{FM-Based Methods for Embodied Navigation}

\begin{figure*}[h!]
    \centering
    \includegraphics[width=0.9\linewidth]{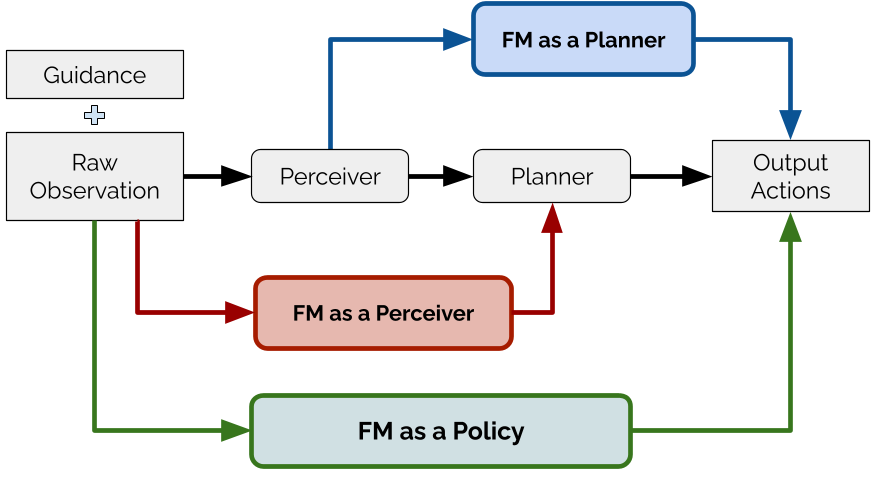}
    \caption{\textbf{Taxonomy}: We consider $3$ classes of FM-based methods for Embodied Navigation. The gray boxes indicate the classical pipeline. 1) FM as a Perceiver (in red) takes the input observation and instruction and produces semantic groundings to be used by a planner. 2) FM as a Planner (in blue) uses the outputs of a perception module to produce subgoals or sequences. 3) FM as a policy goes directly from the input to action.}
    \label{fig:overview}
\end{figure*}

Figure \ref{fig:overview} presents our taxonomy. The gray boxes represent the classical pipeline for reference. We categorize FMs in the following manner:

\begin{itemize}
    \item \textbf{FM as a Perceiver}: In this setting, we consider off-the-shelf FMs trained for vision or language tasks, which find their utility as part of a larger robotics system. The output of the FM is then passed onto a lower level system for planning and control. 
    \item \textbf{FM as a Task Planner}: In this setting, the FM replaces the planner to output a sequence of subgoals for the agent to achieve. These goals are then executed by a separate planner.
    \item \textbf{FM as Policy}: In this setting, the FM directly goes from perception to action, absorbing perception, planning and control into a single model.

\end{itemize}

\noindent\textbf{FM as a Perceiver}: This is represented by the red elements in the taxonomy figure.
FMs can be used as off-the-shelf visual-semantic modules that translate raw observations into
higher-level scene descriptions consumed by a downstream control system.
LGX \cite{lgx} uses an LLM (GPT-3) to perform commonsense reasoning on observations and processed by a visual recognition module (via captions or object detections). The LLM output then drives the navigation policy.
LM-Nav \cite{shah2022lmnav} similarly uses GPT-3 for extracting textual landmarks from free-form instructions, CLIP for grounding those landmarks in a topological map of the environment, and a visual navigation model for execution enabling long-horizon instruction-following. Both these works do not use any form of fine-tuning, and rely only purely on the FM's reasoning capabilities.
CLIP-Nav \cite{clipnav} uses CLIP, a VLM, as a perceiver to directly obtain grounding scores between the target instruction and the scene image to drive a navigation policy.
In a similar vein, VLFM \cite{yokoyama2024vlfm} leverages a frozen VLM to
score candidate frontier regions with language-grounded value maps derived from
the agent's RGB observations, guiding a frontier-based exploration strategy
toward the target object category. 
Both these methods rely on the VLMs spatial reasoning capabilities to achieve superior zero-shot navigation performance, without any finetuning.

\noindent\textbf{FM as a Task Planner}: The is indicated by the blue elements in the taxonomy figure. When used as task planners, FMs decompose high-level navigation
goals into sequences of sub-goals that are handed off to a pre-existing
low-level controller for execution.
The FM here can be an LLM like in LLM-Planner \cite{llmplan1} or as zero-shot planners \cite{llmplan2, llmplan5}, which use it to describe a step-by-step plan for the agent to execute. At test time, it generates a sequence of high level actions like \textit{go to the potato}, which is then grounded in the scene by a low level controller. Another example of this a similar approach is in SayCan \cite{saycan}, which uses environment affordances to determine if the plan produced by the LLM is feasible or not. FM-Planner \cite{llmplan3} uses this to determine which FM is best for drone flights given a set of constraints. ET‑Plan‑Bench \cite{llmplan4} provides a benchmark for FM-based planners by testing various combinations of LLMs/VLMs for task success. This style of usage ties in to an \textit{agentic} framework, where the goal is to identify which subset of combinations can help with task success.

\noindent\textbf{FM as Policy}:
The green arrows in the overview figure indicate this category. Here, we consider an FM that directly translates raw perception data and instructions into actions for navigation. These approaches either train an end-to-end policy entirely using robot data from scratch, or finetune a FM that has been trained on internet data with task-specific robot data.
An example of the former is RT-1 \cite{brohan2023rt1}, which trains a transformer model from scratch using robot data gathered from multiple sources. An example of the latter is RT-2 \cite{zitkovich2023rt2}, which use powerful vision and language encoders trained on internet-scale data to obtain latent representations which are trained to produce robot actions.
Works like GNM \cite{shah2023gnm} and uses the RT-2 principle to train a general model for navigation using data gathered from $6$ different datasets.

\noindent \textbf{Takeaway}:
FM-based approaches for embodied navigation have improved task performance across all 3 categories. However, each category faces unique limitations. FM as a perceiver has high inference latency due to the large compute and poor spatial reasoning due to the lack of embodied training data. FM as a planner often faces issues with executing hallucinated plans that do not make sense in that environment. Further, if the low level controller executed the plan wrong, replanning becomes expensive and can lead to feedback loops. Finally, FM as a policy has massive data gathering requirements to train the reliable end-to-end model. Further, there are risks involved with brittle performance on out-of-distribution scenarios. Even with small changes to the input camera angle, model outputs break entirely \cite{brittle}. We further discuss these challenges in the following section.

\section{Challenges with Deploying FM-Based Embodied Agents}

Here, we outline a few key challenges associated with \textit{training} and \textit{inferencing} FMs on Embodied Agents.


\noindent With respect to \textbf{training}:-
\begin{enumerate}
\item \textbf{Data Scarcity}:  While FMs have achieved great success in the domains of vision, language and audio, large scale data for these domains is available in abundance from the internet for training. Robot data on the other hand is scarce by virtue of there being far fewer robot devices deployed in the real-world, compared to other forms of technology like smartphones. Data for modalities like vision, audio or text has exponentially grown since smartphones became popular; no such technological equivalent exists yet for robotics. As such, capturing valuable real-world data itself becomes a major hurdle for robotics.
\item \textbf{Simulator Training Bias}: Simulators aim to address the problem of data scarcity by creating a representative imitation of the real-world process. However, using simulators to train robot FMs gives rise to other issues with respect to their physical accuracy, realism and perhaps more critically, training \textit{bias}. A model trained with simulator data that does not fully represent the real-world is prone to bias in out-of-distribution cases. Such bias becomes a major issue when the model is transferred to the real-world and is placed  in a situation that it has never seen before in simulation. This can already be seen on modern LLM based chatbots, which often carry over the biases of their human-assisted post-training in their responses.
\end{enumerate}

\noindent With respect to \textbf{inference}:-
\begin{enumerate}
\item \textbf{Large Model Size}: FMs are large models ($>10B$ parameters) that require massive compute. They are often too large for the on-device compute available on several commercially available robot platforms such as the Hello Robot Stretch or the Amazon Astro. Smaller FMs exist ($<10B$ parameters), but their performance is usually far worse than the larger ones \cite{llmscaling}.
\item \textbf{Limited Context Length}:
FMs make decisions based on the context that they have been provided to as input. The size of this context is often limited, and this is a fundamental problem across all Foundation Models. Especially for Embodied tasks which involve sequential decision making by definition, determining what context to keep and what to discard for better on-the-go decisions is of vital importance.
\end{enumerate}

\noindent In the following two sections, we present two approaches that aim to address the training and inference problems discussed above respectively.

\subsection{Personalization: Modelling Portable Targets}

In this work, we first aim to address the FM \textit{training} issues of bias and data scarcity. Instead of training or finetuning an FM, which is computationally expensive, we aim to do this in a zero-shot setting by introducing a `\textit{real2sim}' approach for faithfully representing the environment, and priming the FM with personalized data about the scene.

\begin{table}[b!]
    \centering
    \small
    \begin{tabular}{lccc}
    \textbf{Scenarios} & \textbf{Fixed Rooms} & \textbf{Fixed Paths} & \textbf{Entropy}\\
    \hline
    Static & \greencheck & N/A & None\\
    \hline
    Random & \redx & \redx & High\\ 
    \hline
    Semi-Routine & \greencheck & \redx & Medium\\
    \hline
    Fully-Routine & \greencheck & \greencheck & Low\\
    \hline
    \end{tabular}
     \caption{\textbf{Object Transit Scenarios}: Portable targets transition in a topological graph under structured scenarios inspired by human habits. In the random case, the portable objects can move to any room at any time during each episode. In the routine cases, the rooms that the target objects can travel to are fixed. }
    \label{tab:DOMs}
\end{table}

\begin{figure}[h!]
    \centering
    \includegraphics[width=0.8\linewidth]{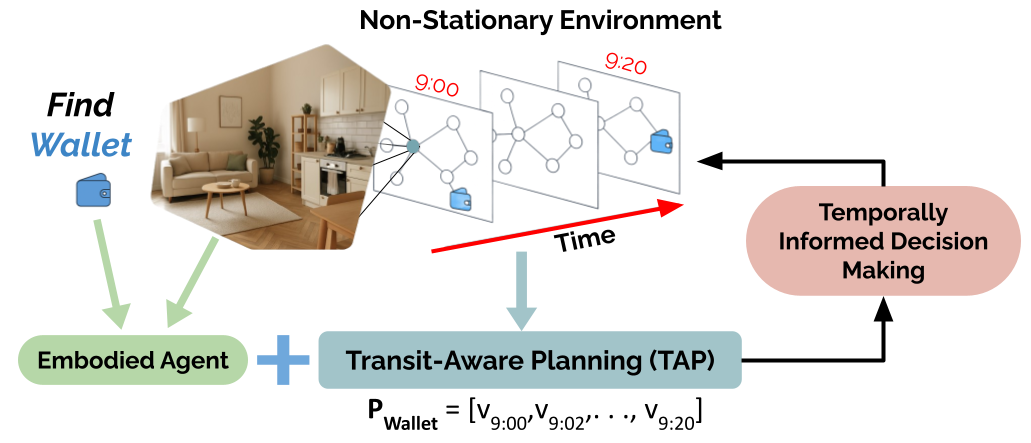}
    \caption{\textbf{Finding Portable Objects}: Objects at home move over time. Modelling these movements in simulation would help with personalized inference during deployment. In this figure, an embodied agent is tasked with finding a wallet that changes positions between 9:00 and 9:20. $P_{\text{wallet}}$ represents the transit route of the wallet. The Transit-Aware Planner (TAP) is learns routines on `real2sim' data which primes the FM-based navigation agents with these routines for temporally informed decision-making.}
    \label{fig:dynamic}
\end{figure}

\begin{figure*}[h!]
    \centering
    \includegraphics[width=0.8\linewidth]{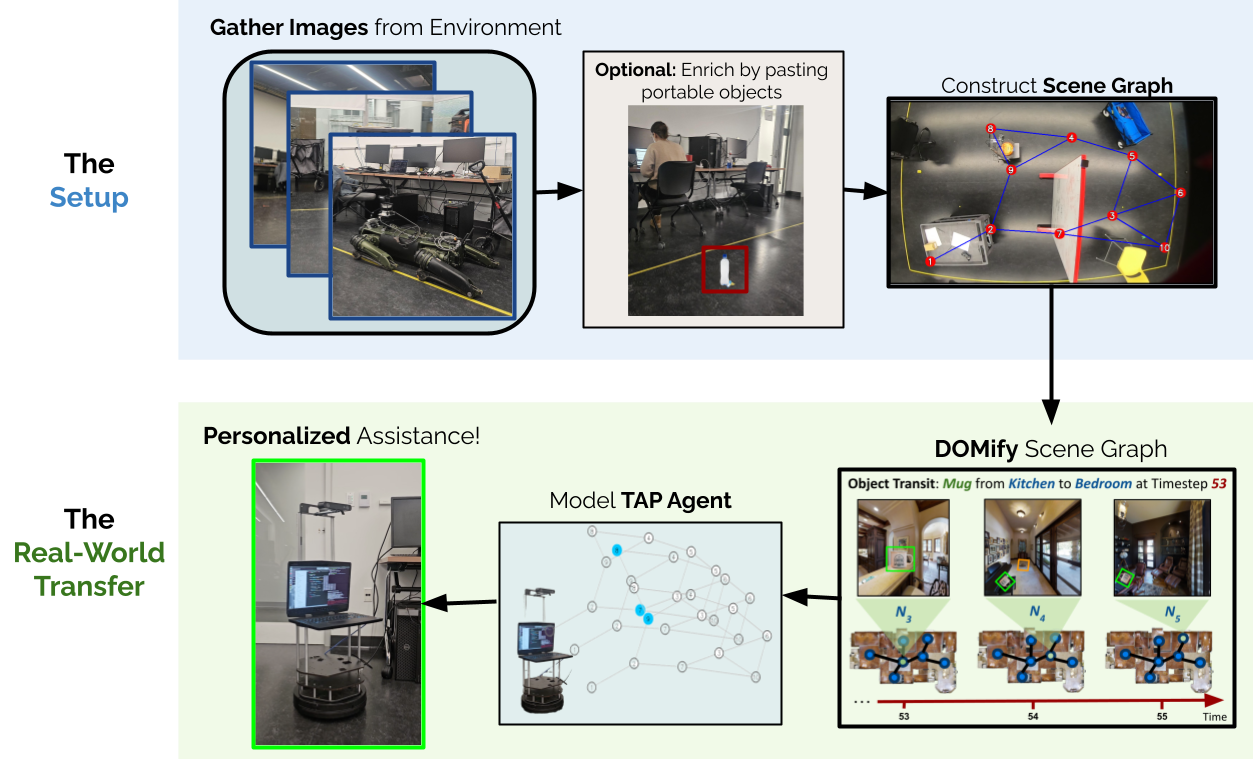}
    \caption{\textbf{Real-world Experiment}: We create a simulation of our lab environment and define DOMs for priming FM-based TAP agents. 
    The first phase involves gathering images from the scene and constructing a topological scene graph (in blue). The graph is then converted to a Dynamic Object Map (DOM) that models the object dynamics, which is then used to train a transit model $T$ for the transit-aware agent (TAP). Finally, we transfer this to the real-world for personalized assistance. Please refer to the video attached for more visuals. Table \ref{tab:rw-stats} presents our results in this environment.
    }
    \label{fig:real-world}
\end{figure*}

\noindent\textbf{Problem Setting}:
Dynamism in the real world can be both real-time and \textbf{non-real-time}. The former is the more popular problem to tackle in the motion planning community, with a robot performing tasks like collision avoidance or seeking-following, in an environment with moving objects (like walking people). The goal is usually to avoid objects while reaching a target. 

\noindent However, motion planning must also take into consideration \textit{non-real-time} dynamics, i.e., environment elements that change positions \textit{over time}, and not in real-time. For example, a coffee mug could be in the kitchen during the day, in the living room in the afternoon, and back to the kitchen later at night. This type of dynamism with objects that move over time is far less discussed in embodied navigation literature, and requires capturing \textit{temporal} information about object placement in scenes. Knowing the object placement habits of humans in the scene would enable more personalized decision-making during deployment in household/office environments.

\begin{table}[h!]
    \centering
    \begin{tabular}{c|c|c|c}
         & Random $\Lambda$ (\%)& Semi-Routine $\Lambda$ (\%)& Routine $\Lambda$ (\%)\\       
        LLM & $37.5$ & $45.0$ & $57.5$ \\
        TAP-LLM & $\textbf{50.0}$ & $\textbf{65.0}$ & $\textbf{80.0}$
    \end{tabular}
    \caption{\textbf{Real World SR}: We report average success rates for finding portable objects in our lab environment over $50$ episodes. Note the improved performance of the TAP-LLM approach over a vanilla LLM scheme (\cite{lgx}). Transit awareness helps improve performance in the routine cases, where paths have structure.}
    \label{tab:rw-stats}
\end{table}

\begin{figure}[b!]
    \centering
    \includegraphics[width=0.8\linewidth]{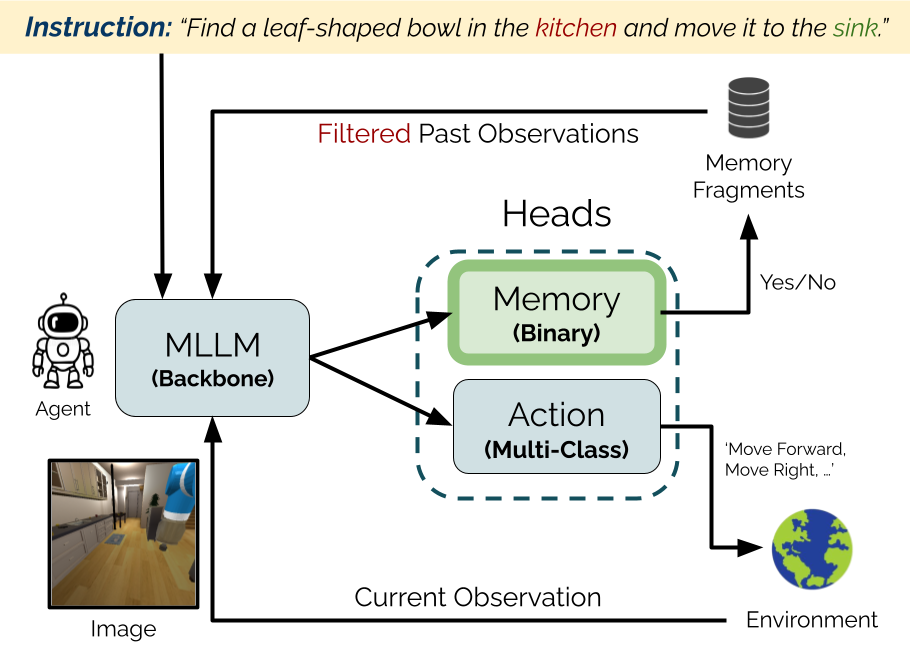}
    \caption{\textbf{MemCtrl ($\mu$)}:
    We present MemCtrl, a novel memory filtering scheme to improve decision making performance on \textbf{small} MLLMs tackling embodied tasks.
    Our approach proposes a \textit{trainable memory head} (green box) that learns to actively filter out redundant observations on-the-go. This form of \textit{active} filtering aims to alleviate issues inefficient retrieval from stored observations, while also enabling scalable deployment on FMs as a detachable memory head.
}
\label{fig:memctrl_overview}
\end{figure}

\noindent We pose this as a `portable' ObjectNav \cite{objectnavrevisited} problem in non-stationary environments, where the agent must 1) model human object-placement habits in the real-world, and 2) utilize this in its planning policy for temporally-aware decision making. An overview of this problem setup is in Figure \ref{fig:dynamic}.

\noindent\textbf{Solution}: Our solution involves two parts. We first model object placements that could occur in the real world as Dynamic Object Maps (DOMs). DOMs convert topological scene graphs constructed from real-world environments into dynamic node-attributed graphs, with objects moving over time. DOMs aim to capture various possible object transit scenarios (Table \ref{tab:DOMs}). We then learn the transit behavior to obtain a human habit model $T$, which is used to prime the FM with transit information. Figure \ref{fig:real-world} illustrates our real-world experiment, and the attached video has more visuals. 

\begin{figure}[b!]
    \centering
    \includegraphics[width=0.7\linewidth]{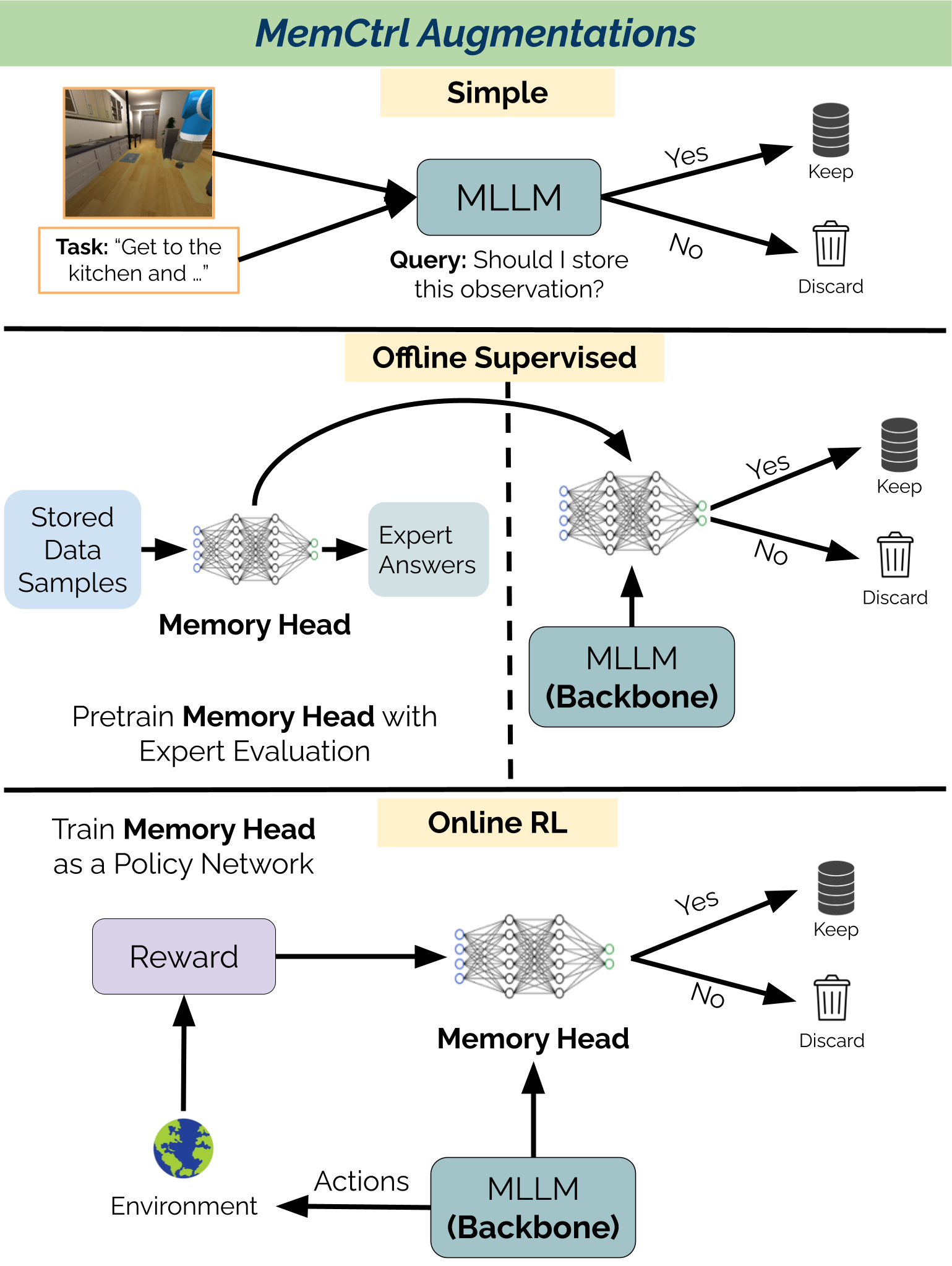}
    \caption{\textbf{MemCtrl Variants}: We experiment with $3$ variants. The simple case acts as a non-trained baseline, where the MLLM is directly queried about storage. In the offline supervised case, $\mu$ is first pretrained using expert answers from a high performing, expert MLLM (GPT-4o here). This trained binary classifier then acts as a head on top of the MLLM backbone. In the Online RL case, we train the memory head online as a policy network. We use a sparse reward on task success and a dense reward on action success. Note that MemCtrl is trained as a detachable head that takes the latent MLLM embeddings as input.}
    \label{fig:memctrl_approach}
\end{figure}

\subsection{Memory Efficient FMs}


\begin{figure}[t!]
    \centering
    \includegraphics[width=0.75\linewidth]{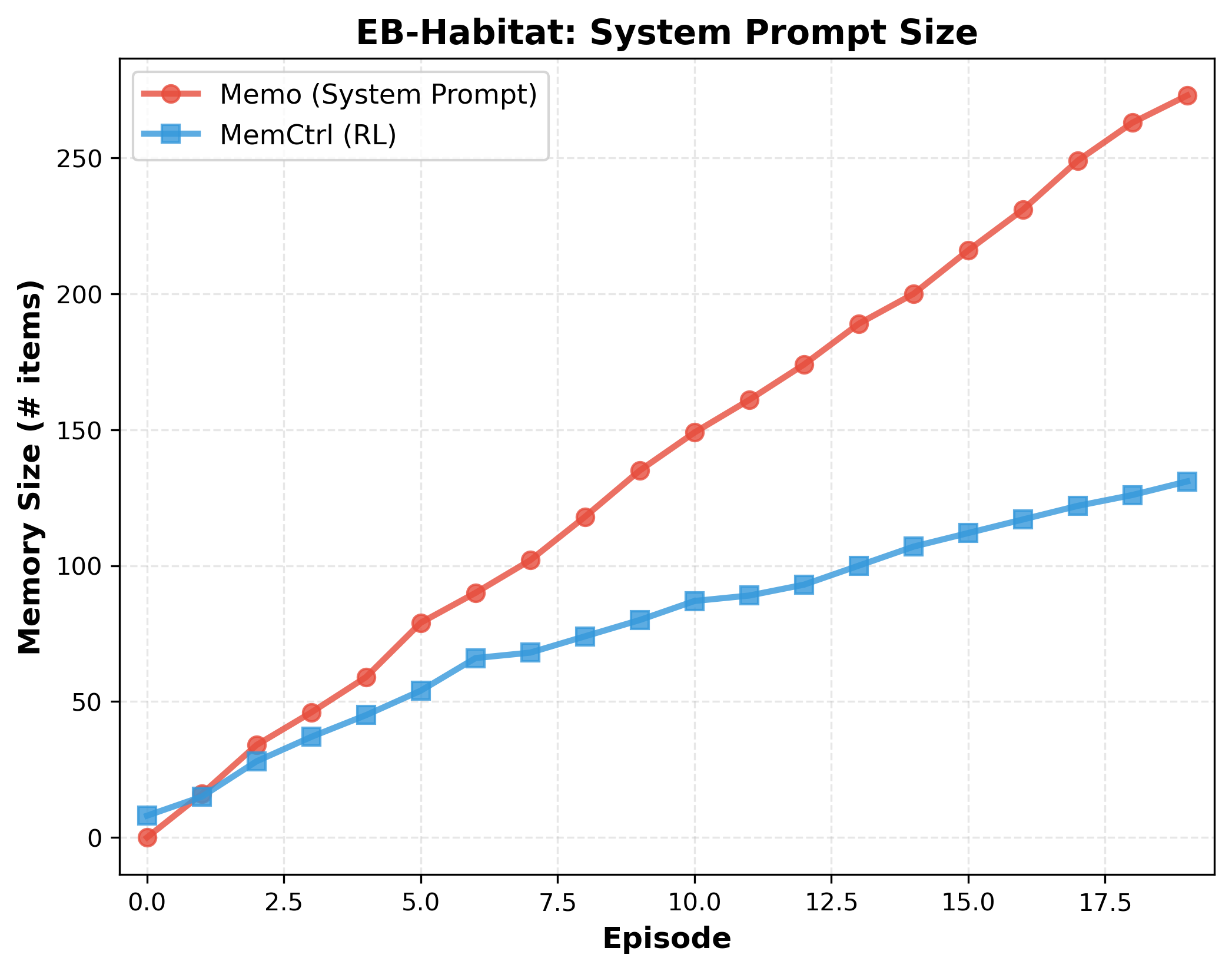}
    \caption{\textbf{Context Growth}: MemCtrl (in blue) shows for lesser memory usage and better performance over a baseline Memo-style (in red) approach that stores the entire history as context. These results are on the EB-Habitat environment with Qwen2.5-VL-7B-Ins augmented with an offline supervised $\mu$, using GPT-4o data.}
    \label{fig:memctrl_memory}
\end{figure}

\begin{table*}[h!]
\centering
\resizebox{\linewidth}{!}{
\begin{tabular}{l|cccccc|cccccc}
\toprule
\textbf{Model} &
\multicolumn{6}{c|}{\textbf{EB-ALFRED}} &
\multicolumn{6}{c}{\textbf{EB-Habitat}} \\
 & \textbf{Avg} & \textbf{Base} & \textbf{Common} & \textbf{Complex} &
   \textbf{Spatial} & \textbf{Long} &
   \textbf{Avg} & \textbf{Base} & \textbf{Common} & \textbf{Complex} &
   \textbf{Spatial} & \textbf{Long} \\
\midrule
Qwen2.5-VL-7B-Ins & $5.2$ & $10$ & $8$ & $6$ & $1$ & $2$ & $15.2$ & $32$ & $2$ & $26$ & $14$ & $2$ \\
Qwen2.5-VL-7B-Ins + $\mu_{\textbf{Simple}}$ & $7.4$ & $11 \pm 3$ & $9 \pm 1$ & $8 \pm 1$ & $2 \pm 1$ & $7 \pm 1$ & $19.0$ & $\underline{39} \pm 3$ & $\underline{5} \pm 1$ & $31 \pm 2$ & $14 \pm 1$ & $6 \pm 2$ \\
Qwen2.5-VL-7B-Ins + $\mu_{\textbf{Offline Sup.}}$ & $8.8$ & $\underline{16} \pm 2$ & $10 \pm 1$ & $7 \pm 3$ & $2 \pm 0$ & $\underline{9} \pm 3$ & $\textbf{21.0}$ & $38 \pm 2$ & $5 \pm 0$ & $30 \pm 1$ & $\underline{18} \pm 1$ & $\underline{14} \pm 1$ \\
Qwen2.5-VL-7B-Ins + $\mu_{\textbf{Online RL}}$ & $\textbf{9.8}$ & $10 \pm 2$ & $\underline{13} \pm 1$ & $\underline{15} \pm 4$ & $\underline{3} \pm 1$ & $8 \pm 3$ & $18.4$ & $32 \pm 3$ & $3 \pm 0$ & $\underline{32} \pm 3$ & $14 \pm 1$ & $11 \pm 1$ \\
\bottomrule
\end{tabular}
}
\caption{\textbf{MemCtrl Results}: We augment \textit{Qwen2.5-VL-7B-Ins} with the 3 variations of MemCtrl on 5 subsets of EB-ALFRED and EB-Habitat \cite{yang2025embodiedbench} across 3 trials. Note the improve performance overall of adding the memory head $\mu$. In particular, we note superior performance on long context and complex instructions, which tend to be long horizon where memory is particularly useful.
}
\label{tab:memctrl_results}
\end{table*}

\begin{figure*}[h!]
    \centering
    \includegraphics[width=\linewidth]{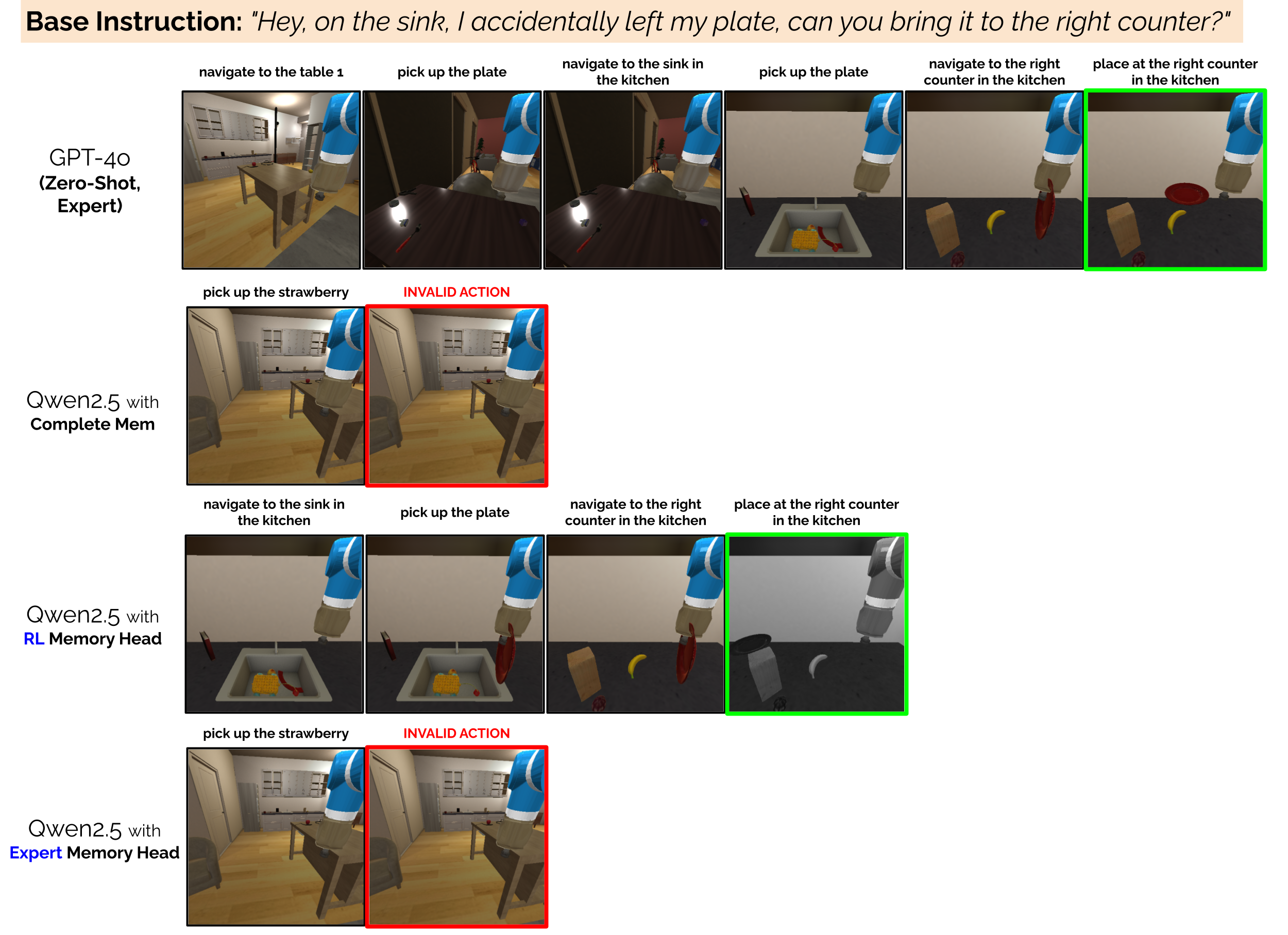}
    \caption{\textbf{Base Performance on EB-Habitat}: Here, we compare the performance of GPT-4o vs Qwen2.5-VL-7B-Ins, with various memory augmentations. The green box around the image indicates successful task completion, while the red box indicates otherwise. The images are agent observations.
    Note GPT-4o takes more steps to complete the task. On the other hand, $\mu$ boosts the performance of a significantly weaker model, and in this scenario, even doing it quicker with $\mu_{RL}$. Grayed out images indicate discarded memories.}
    \label{fig:base}
\end{figure*}

\begin{figure*}[h!]
    \centering
    \includegraphics[width=\linewidth]{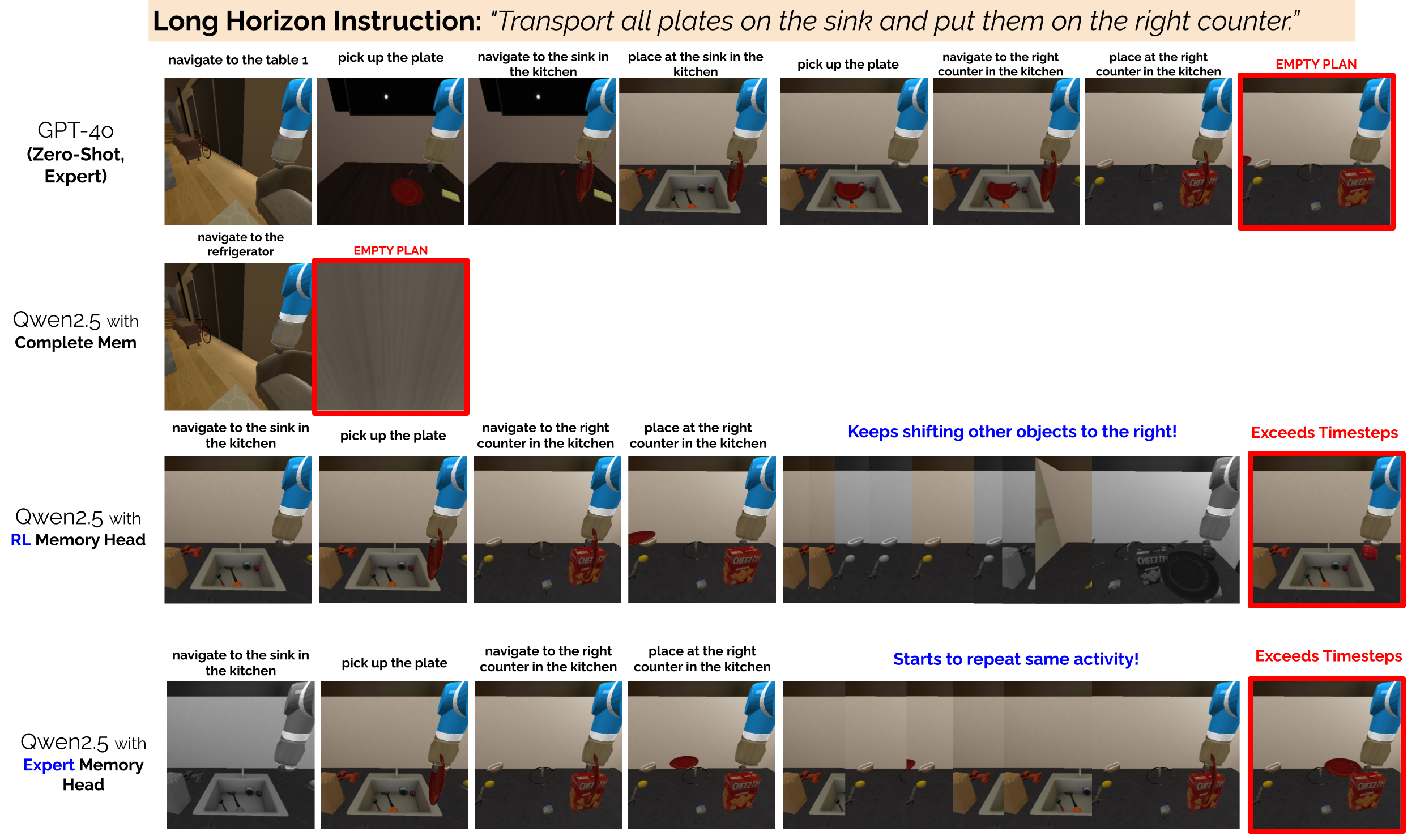}
    \caption{\textbf{Long Horizon performance on EB-Habitat}: We notice that on long horizon tasks, the expert tends to end the task early by hastily assuming that it is done (finishing after placing \textit{one} plate instead of \textit{all} plates). Memory heads highlight unique performance improvements, with $\mu_{\text{RL}}$ exhibiting a more \textbf{exploratory} nature by continuing to place \textit{new} objects at the right counter, and $\mu_{\text{Expert}}$ being more \textbf{exploitative} by repeating the same activity over and over, with a single plate. Investigating this is part of a future study. \textit{Note}: Grayed out images indicate discarded memories.}
    \label{fig:long}
\end{figure*}

\noindent To help mitigate FM issues of limited context length and deployment on low-compute robots, we propose MemCtrl, an active memory head that learns to filter out useful observations on the go. This approach is highlighted in Figure \ref{fig:memctrl_overview}. 

\noindent Memory Heads ($\mu$) have the following features:-

\begin{enumerate}
    \item \textbf{Active Memory Filtering:} 
    $\mu$'s trained on top of a frozen MLLM backbone to actively filter observations to determine which to keep and which to discard in memory. Unlike prior retrieval-based work involving filtering large observational data offline, $\mu$ enables the MLLM to engage in real-time filtering, which is particularly useful in memory-constrained settings involving small models. 

\end{enumerate}

\noindent We experiment with $3$ different types of MemCtrl agents on low parameter MLLMs on the EmbodiedBench benchmark. These are highlighted in \ref{fig:memctrl_approach}. Figure \ref{fig:memctrl_memory} compares the MemCtrl agent's memory usage over time against a standard FM-based agent (Memo). Note the reduced memory footprint on a small parameter FM with an improved performance.

Table \ref{tab:memctrl_results} contains our results. We note an improved performance of $6\%$ on average with nearly $20\%$ on long horizon instructions. We make the following key observations:
\begin{itemize}
    \item \textbf{Improved Performance}: Adding any variant of $\mu$ improves performance across all settings. This shows the effectiveness of active memory filtering. 
    In particular, we observe improvements on long form instructions, and this reflects the influence of better memory management on such tasks.

    We also observe an overall improvement in the task performance on complex instructions, where the instructions are not just long, but also contain irrelevant information. For instance, the following is the difference between a base and complex instruction:- 
\begin{quote}
    \textbf{Base}: ``Move one of the pear items to the indicated sofa.''
\end{quote}

\begin{quote}
    \textbf{Complex}: ``When you find the fridge door open, go ahead and move an bowl to the sofa; otherwise, transport an hammer to the sofa.''
\end{quote}

The agent is expected to finish these tasks in a fixed set of timesteps, and over time, gathers more and more information about the environment as potential context for determining its next action. The base query here is fairly simple, requiring to track just a single object (`pear'). In contrast, the complex query not only has multiple objects to track (`fridge, sofa, bowl, hammer'), but is also sophisticated in its framing, requiring better reasoning. While more context would help with better reasoning, it also leads to more redundant information storage, which a trained memory head can help actively filter.

    \item \textbf{ALFRED vs Habitat}: We note the a general improvement in performance on Habitat versus ALFRED tasks. Habitat has more navigation heavy tasks, while ALFED is more manipulation heavy. This result suggests that memory heads are particularly useful on navigation tasks that require long-horizon planning.
    \item \textbf{Qualitative Results}: Figure \ref{fig:base} and \ref{fig:long} show a qualitative analysis of the memory heads on a long-horizon task from EB-Habitat. 
    Figure \ref{fig:base} deals with a base case, with a simple task, where the $\mu$ boosts performance. In Figure \ref{fig:long} however, which is a long-horizon task, the baseline method GPT-4o fails to complete it. It navigates to the table to pick up a plate, but ends the task too early. We observe that both $\mu$ agents continue the activity, but in different ways. $\mu_{RL}$ is more exploratory by finding new objects to place each time, while $\mu_{Expert}$ (offline supervised) is more exploitative, in repeating the same activity over and over. Analysing this unique behaviour is part of future work.
\end{itemize}


\section{Conclusion \& Future Directions} 

We present two novel contributions towards enabling the deployment of Foundation Models (FM) based Embodied Agents in the real-world. The first contribution is aimed at mitigating training bias for a portable object finding task, which we address by priming the FM with learned behaviours of dynamic objects. The second contribution tackles the inference time issues of limited context length and large model size, by introducing a novel `memory head' architecture, MemCtrl. MemCtrl improves the performance of low-parameter FMs with lower context size. In the following subsections, we discuss some potential future directions for research, and our stance on the deployability of FM-based embodied agents.

\noindent On \textit{Day-One} Unboxing:
Robot agents deployed in human-centric environments such as households or office spaces are expected to reliably perform a wide variety of tasks right out of the box. This objective is non-trivial, as beyond hardware constraints like poor tactile sensing and form factor, different environments could have a large variance in terms of perceptual semantics and target instructions. Prior to deploying such agents, it becomes important to have a reasonable estimate of how well they may perform in such unseen environments. We pose this as the day-one unboxing problem and ask,

\begin{quote}
    \textit{What are the minimum set of tasks that an agent must successfully perform right after unboxing for it to be considered reasonably intelligent?}
\end{quote}

\noindent Framing this as a day-one unboxing problem attempts to calibrate the agent, allowing the robot manufacturer to have an estimate on how close the unboxed environment was to the data the FM was trained on.
\noindent Analogous to smartphones which successfully perform a variety of tasks right out of the box, we hypothesize that for an Embodied agent to be considered useful, it too must be able to perform a variety of tasks successfully, right out of the box. We pose two research questions along these lines which need to be answered before deployment:- \\
\noindent \textit{RQ1: What core functions must the FM on a robot be able to reliably achieve before deployment?}\\
\noindent \textit{RQ2: What functions \textbf{must not} be available beforehand, and how must the agent use context from the scene to achieve its goals?}

\noindent RQ1 is in relation to functions that a deployable agent must be able to achieve right after unboxing such that it forms a positive impression upon the user. This could include primitive tasks like pick and place, or exploratory navigation.

\noindent RQ2 is more in regards to what functions the agent must not be trained for, but rather pick up after unboxing. We postulate that rather than training an FM to learn every possible task (i.e., fully general behaviour), it must focus on a few critical skills beyond which it must look for context in the scene to achieve zero-shot success. Overfitting on every task is not beneficial at all, as it might catastrophically fail in novel settings (\textit{putting away groceries}, for instance). We believe agents would benefit from a system that encapsulates the FM for decision making, rather than the FM alone as a policy.

\bibliographystyle{bibtex/splncs03}
\bibliography{refs}

\end{document}